\documentclass[conference]{IEEEtran}
\IEEEoverridecommandlockouts
\usepackage{cite}
\usepackage{amsmath,amssymb,amsfonts}
\usepackage{algorithmic}
\usepackage{graphicx}
\usepackage{textcomp}
\usepackage{xcolor}
\usepackage{booktabs}
\usepackage{array, makecell}
\usepackage{caption}
\usepackage{subcaption}
\usepackage{dsfont}
\usepackage{tabularx}
\usepackage{multirow}
\usepackage{bbm}
\usepackage{hyperref}
\def\BibTeX{{\rm B\kern-.05em{\sc i\kern-.025em b}\kern-.08em
    T\kern-.1667em\lower.7ex\hbox{E}\kern-.125emX}}

\newcommand{\eg}{\textit{e}.\textit{g}.}
\newcommand{\ie}{\textit{i}.\textit{e}.}
\newcommand{\cf}{\textit{c}.\textit{f}.}
\begin{document}

\title{BILLNET: A Binarized Conv3D-LSTM Network with Logic-gated residual architecture for hardware-efficient video inference}

\author{\IEEEauthorblockN{Van Thien Nguyen, William Guicquero and Gilles Sicard}
Smart Integrated Circuits for Imaging Laboratory, CEA-LETI \\
F-38000, Grenoble, France. [Email: vanthien.nguyen@cea.fr]}

\makeatletter
\def\ps@IEEEtitlepagestyle{
  \def\@oddfoot{\mycopyrightnotice}
  \def\@evenfoot{}
}
\def\mycopyrightnotice{
  {\footnotesize
  \begin{minipage}{\textwidth}
  \centering
  Copyright~\copyright~2022 IEEE. Personal use of this material is permitted. However, permission to use this material \\ 
  for any other purposes must be obtained from the IEEE by sending an email to pubs-permissions@ieee.org. DOI: \href{https://ieeexplore.ieee.org/document/9919206}{10.1109/SiPS55645.2022.9919206}
  \end{minipage}
  }
}

\maketitle

\begin{abstract}
Long Short-Term Memory (LSTM) and 3D convolution (Conv3D) show impressive results for many video-based applications but require large memory and intensive computing. Motivated by recent works on hardware-algorithmic co-design towards efficient inference, we propose a compact binarized Conv3D-LSTM model architecture called BILLNET, compatible with a highly resource-constrained hardware. Firstly, BILLNET proposes to factorize the costly standard Conv3D by two pointwise convolutions with a grouped convolution in-between. Secondly, BILLNET enables binarized weights and activations via a MUX-OR-gated residual architecture. Finally, to efficiently train BILLNET, we propose a multi-stage training strategy enabling to fully quantize LSTM layers. Results on Jester dataset show that our method can obtain high accuracy with extremely low memory and computational budgets compared to existing Conv3D resource-efficient models.                            
\end{abstract}


\section{Introduction}
Video recognition has recently drawn substantial attention due to the success of several Deep Neural Networks (DNNs) \cite{GSN}, \cite{TSN}, \cite{Gating_revisited} and the increasing number of large-scale video datasets \cite{Kinetics}, \cite{Utube8M}, \cite{Jester}. Compared to other tasks like image classification that only relies on spatial data, video recognition is much more complex since it also requires extracting the underlying temporal features in the time direction. Among existing model architectures for spatio-temporal pattern recognition, Conv3D \cite{C3D} and Recurrent Neural Networks (RNNs), \eg \,LSTM \cite{LSTM}, have demonstrated to be relevant for learning latent spatio-temporal representations. However, these model components exhibit hardware-related drawbacks such as large memory requirements so as a high computational complexity. For instance, Conv3D expands the convolution kernel to the time direction for capturing local temporal features, therefore increasing both the local memory and computational needs by an order of magnitude compared to Conv2D. On the other hands, LSTM is computationally expensive because of its stateful nature, \ie, computing the current features taking into account previous states. Consequently, designing a hardware-compliant Conv3D-LSTM model for embedded inference applications remains a significant challenge.

Motivated by the need for efficient video inference, recent works have been focusing on the design of light-weight architectures \cite{RE3DCNN}, \cite{TinyVN} or hardware-aware network pruning \cite{3D-FPGA-Pruning}. Another approach to accelerate the computation during inference and further reduce the hardware-related costs consists in lowering the bitwidth of model's weights and activations \cite{QNN}. Even though significant works on Quantization Aware Training (QAT) \cite{BNN}, \cite{LSQ} have been done in recent years, they are mostly focused on the quantization of feed-forward Convolutional Neural Networks (CNNs). On the contrary, training fully quantized models that embed recurrent layers like LSTMs remains an important issue. Indeed, quantizing the hidden states of LSTM involves a quantization error that is accumulated throughout the data sequence due to its recurrent nature, hence implying an overall accuracy degradation. It may explain why existing approaches \cite{RNN_low_prec}, \cite{EQRNN} are limited to the use of quantization regarding the LSTM weights while keeping activations in a floating-point representation.      

This work thus aims at demonstrating that a fully-quantized model can also be deployed for video-based inference. To this end, we propose a hardware-compliant Conv3D-LSTM architecture called BILLNET on which binarization techniques are applied to further reduce the model size as well as the computational costs. Our main contributions are then:
\begin{itemize}
\item A compact Binarized Conv3D-LSTM model architecture with a MUX-OR skip connection mechanism,
\item A multi-stage training procedure that provides a fully quantized BILLNET with bitshift normalization (removing additional biases related to batch normalization).
\end{itemize}
\section{Related works}
\subsection{Residual connections for video inference} 
Residual learning such as element-wise addition \cite{ResNet} and attention mechanism \cite{ResAttention}, \cite{SeNet} is firstly introduced in 2D CNNs in order to increase network expressivity, favor feature reuse while easing the back-propagation for deeper models.
Since Conv3D has become a more preferable option for video recognition than its 2D counterpart, it is straightforward that 3D CNNs should adopt residual architecture paradigms like element-wise addition \cite{3DResNet}, \cite{Revisiting3RN} and attention mechanisms \cite{ST_Attention} to improve model performance. However, it is noteworthy mentioning that these skip connection operations are mostly performed using full-precision arithmetic, which results in additional hardware-related costs, especially in the context of fully-quantized models (targeting a dedicated hardware mapping). In our work, we thus introduce a 3D quantized MUX layer with an OR-gated connection that allows integrating both an element-wise additional connection and a channel-wise attention-like mechanism, while keeping a binarized data representation.

\subsection{Efficient 3D CNN architecture.}
Several works have recently proposed alternative architectures to alleviate the parameter-heaviness of Conv3D. \cite{MixedConv}, \cite{RethinkingConv3D_byMixedConv} partly replace Conv3D by 2D convolutions. \cite{P3D} proposes different variants for the 3D residual block by separating $3\times3\times3$ kernels with $1\times3\times3$ spatial convolutions and $3\times1\times1$ temporal convolutions. On the other hand, \cite{TSM} processes the temporal features without parameters and multiply-accumulate (MAC) operations by shifting part of the channels along the time dimension. Finally, \cite{RE3DCNN} converts various well-known resource-efficient 2D CNNs such as MobileNet \cite{MobileNet}, ShuffleNet \cite{ShuffleNet} to 3D CNNs. In our work, BILLNET uses a basic factorization closely related to the one of 3D ResNext \cite{3DResNet}, including 2 pointwise layers and a grouped Conv3D, without nonlinearity (\ie, activation) inserted in-between.   

\subsection{Network quantization.}
Network quantization \cite{QNN} reduces the bitwidth of weights and/or activations. In the most extreme case, Binarized Neural Networks (BNNs) \cite{BNN} restrict both weights and activations to a 1-bit representation using Sign function, this ways reducing the costly full-precision MACs to bitwise operations (\ie, using XNOR gates). Despite tremendous progress during the last few years, there still exists a lack of efforts on model quantization for video inference. Recently, \cite{DynamicQ} adaptively selects the per-frame optimal bitwidth, conditioned on input data. \cite{Bin3DCNN} proposes a binary 3D CNN constraining weight and activation values to $0$ or $1$. Besides, existing approaches \cite{4bQLSTM}, \cite{EQRNN}, \cite{HitNet}, \cite{DCLSTM} mostly focus on the compression of LSTM for language or speech models only. To the best of our knowledge, there is no prior works on fully-quantized LSTM in the context of video inference. Our work tries to fill in this gap by proposing a multi-stage training algorithm to provide a fully-quantized Conv3D-LSTM model.       

\section{BILLNET}
\begin{figure}[htbp]
	\centerline{\includegraphics[scale=0.25]{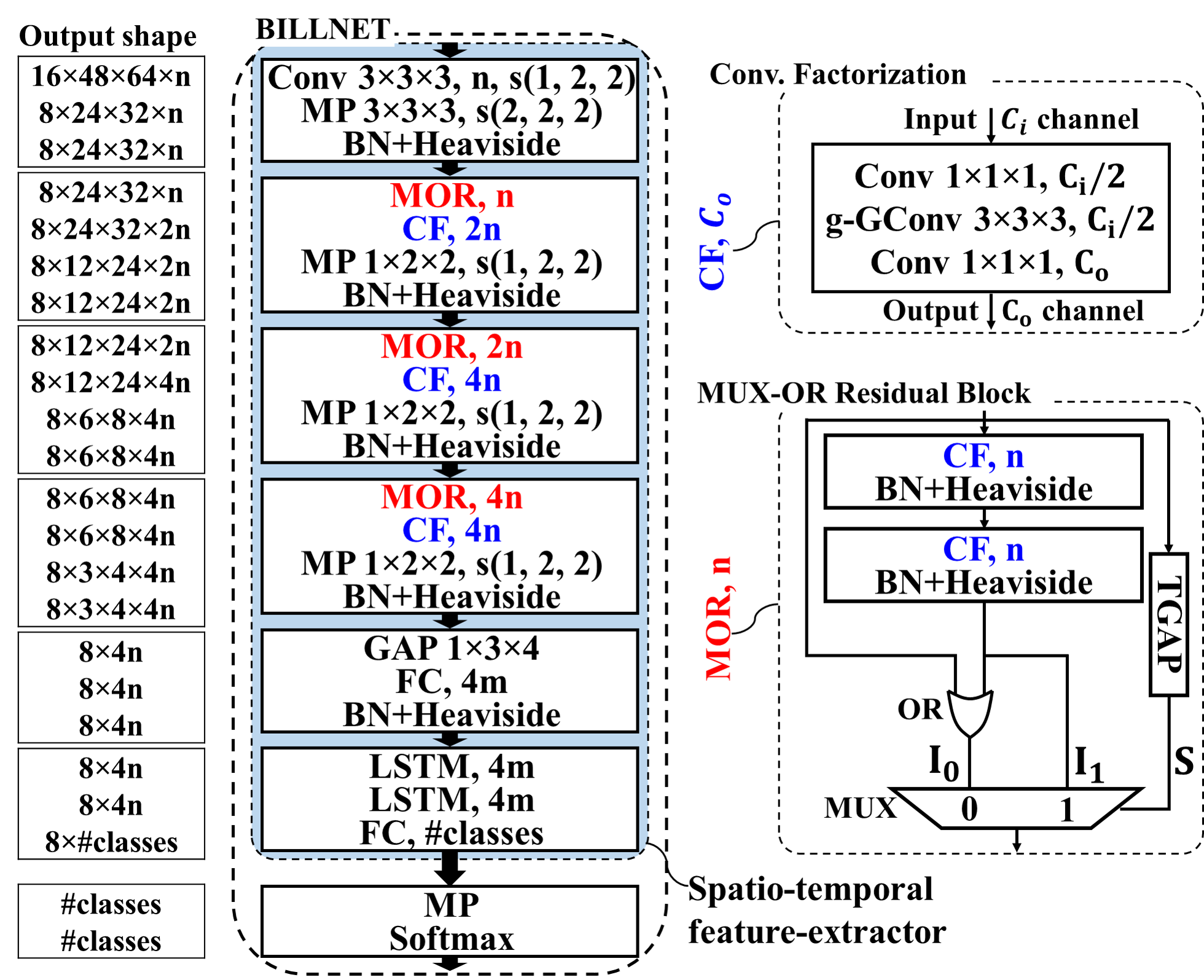}}
	\caption{Top-level architecture description of BILLNET with Convolutional Factorization (\textcolor{blue}{CF}) and MUX-OR Residual (\textcolor{red}{MOR}) Block. Here $n$ is the parameters controlling the number of output feature maps, g-GConv is Grouped Convolution with g groups, MP and GAP stand for Max Pooling and Global Average Pooling.} 
	\label{BILLNET}
\end{figure}

Figure~\ref{BILLNET} depicts the top-level view of BILLNET that involves integer-only MACs and bitwise operations such as 2-input multiplexers and OR gates. This model takes as input a sequence of 16 frames with a spatial resolution of $96\times128$. BILLNET contains a spatio-temporal feature extractor with a Conv3D part to extract spatio-temporal features between adjacent frames, and a LSTM part to keep track longer-term temporal dependencies. In this section, we first focus on the 3D Convolution Factorization (CF), then on the custom 3D MUX-OR Residual (MOR) block, finally on the LSTM weights and activations quantization.   

\subsection{Conv 3D Factorization}
The core building block of BILLNET is a light-weight factorization, namely CF, consisting of 2 pointwise layers (filter size: $1\times1\times1$) and a $g$-grouped convolution (filter size: $3\times3\times3$). Unlike the building block of 3D ResNext in \cite{3DResNet}, there is no nonlinearity (\eg, normalization, activation) between these layers. The number of output channels $C_o$ of each CF is defined by the parameter $n$ (\ie, $C_o \in \{n, 2n, 4n\}$). Note that, the number of channels in low dimension of every CF is set to $C_i/2$.

\subsection{3D MUX-OR Residual (MOR) block} \label{MOR}
\begin{figure}[h]
     \centering
         \begin{subfigure}[b]{0.49\textwidth}
        \includegraphics[width=\textwidth]{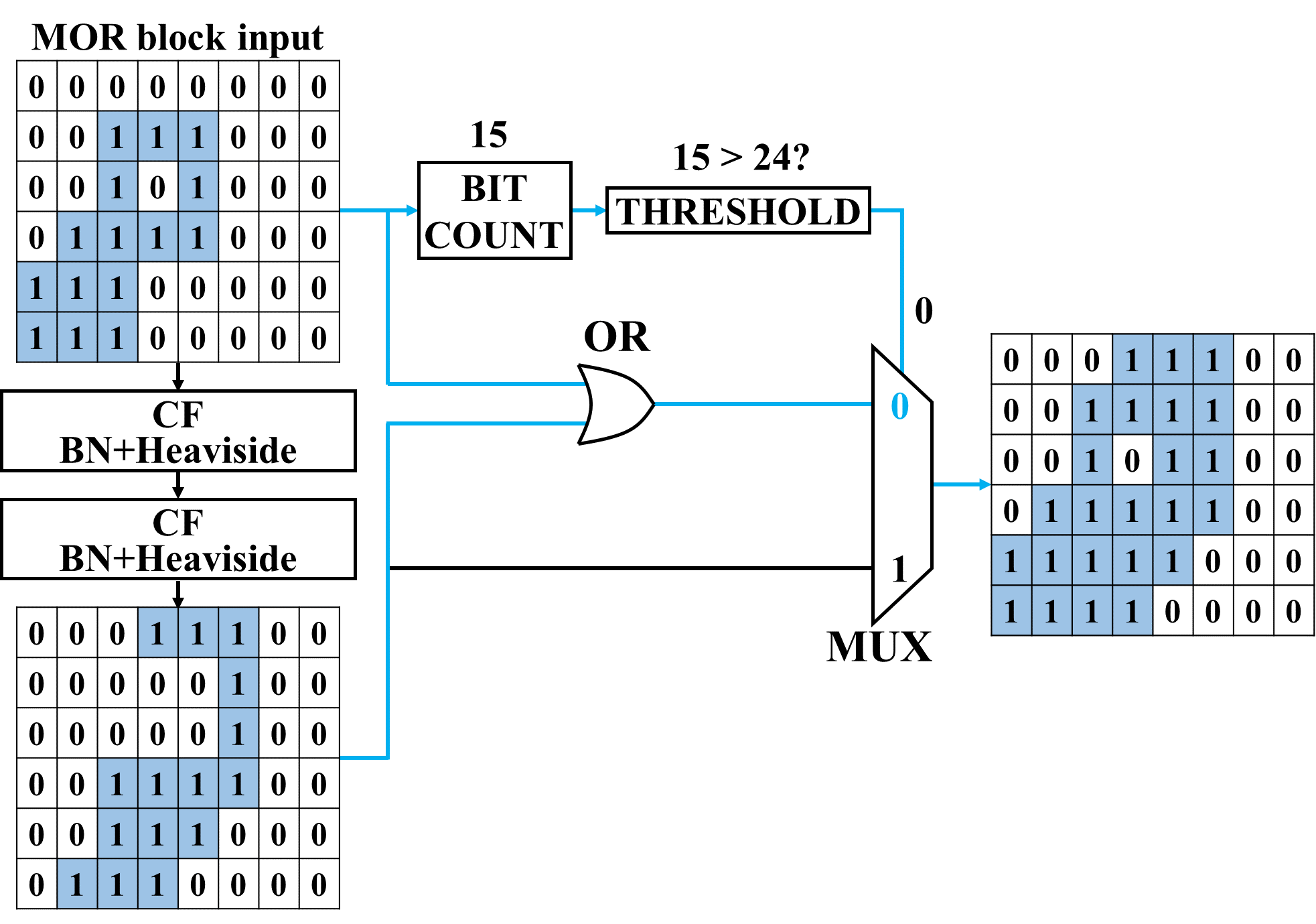}
        \caption{Case of an input dominated by zeros.}
        \label{mux_0}
    \end{subfigure}
    \hspace{-2mm}
    \begin{subfigure}[b]{0.49\textwidth}
        \includegraphics[width=\textwidth]{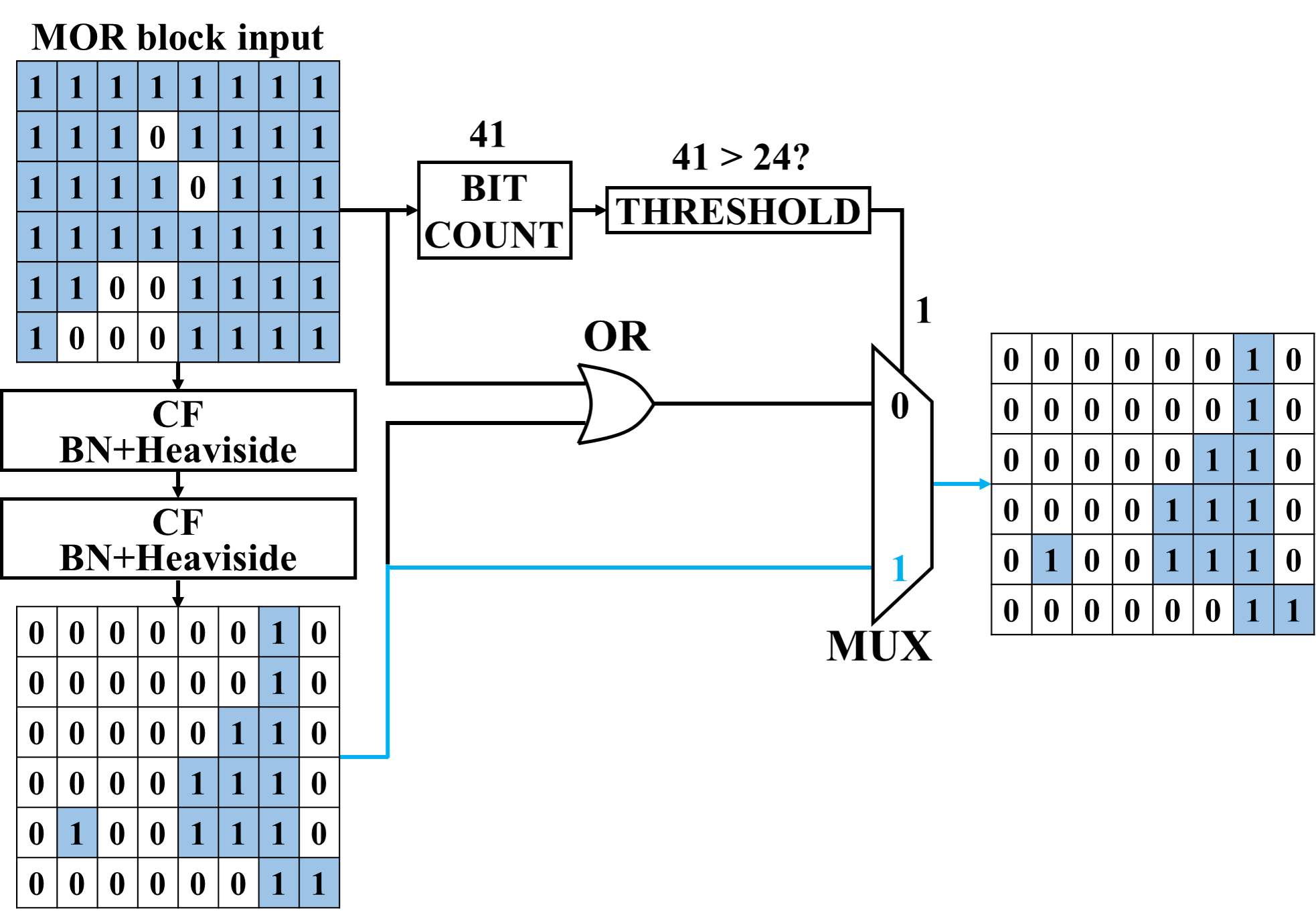}
        \caption{Case of an input dominated by ones.}
        \label{mux_1}
    \end{subfigure}
    \hspace{-2mm}
     \caption{The operation of the channel-wise MUX gate with feature maps extracted during inference of a test sample. The TGAP is implemented by a bitcount followed by an integer-to-integer comparison, where the threshold is equal to one half of the spatial resolution ($\frac{6 \times 8}{2}=24$).}
     \label{MOR_demo}
    \vspace{-3mm}
\end{figure}

In BILLNET, we binarize all the activations using the Heaviside function  $H(x) = \mathds{1}_{\{x > 0\}}$, where $\mathds{1}$ is the indicator function. During the backward pass, the Straight-Through-Estimated gradient (STE\cite{STE}) $\frac{\partial \textrm{H}}{\partial \boldsymbol{x}} = \mathds{1}_{\{|\boldsymbol{x}|\leq 1\}}$ is used. For the sake of genericity, we then define the Clipped Identity: 
\begin{equation}
\textrm{Clip} (y) =  \textrm{max}\left(-1, \textrm{min} \, (1, y) \right)
\label{clipping}
\end{equation}
 with STE gradient $\frac{\partial \textrm{Clip}}{\partial \boldsymbol{y}} = 1$. Concretely, applying this function to the sum of $x_1, x_2$ will obtain the same binary output $\{0, 1\}$ as performing the logical \textbf{OR} operation, \ie, $\textrm{Clip} (x_1 + x_2) = x_1 \lor x_2$. Therefore, we employ the Clipped Identity to keep the data in binary representation. Let us denote $\textbf{I}_0, \textbf{I}_1 \in \mathbb{R}^{T\times h\times w \times n}$ as the output of this OR operation and the second Heaviside where $T, h, w, n$ are the time steps, height, width and number of channels; $\textbf{S} \in \{0, 1\}^{T\times1\times1\times n}$ as the binary control signal. The 2-MUX layer can be described as:
\begin{equation}
\textrm{\textbf{MUX}} (\textrm{\textbf{I}}_0, \textrm{\textbf{I}}_1; \textrm{\textbf{S}}) = \textrm{\textbf{I}}_1 \odot \textrm{\textbf{S}} + \textrm{\textbf{I}}_0 \odot (\textbf{1} - \textrm{\textbf{S}})   
\label{mux}
\end{equation}
where $\odot$ is a channel-wise multiplication. The control signal $\textbf{S}$ embeds a parameter-free channel attention through a Thresholded Global Average Pooling (TGAP). TGAP simply consists of a channel-wise Average Pooling (AP) with filter size and strides of $1\times h\times w$, followed by a binarization $T(x) = \mathds{1}_{\{x > 0.5m\}}$ where $m$ is first set to the maximum of the layer-wise AP's tensor outputs for the full-precision mode, then being replaced by $1$ for the final quantized model version (see section \ref{MSQAT}). When deploying the quantized model, this operation can be implemented via a bit-count followed by an integer-to-integer comparison. This architecture allows the input of each MOR to control the operation of the MUX gate in a channel-wise manner. In details, if the input feature map is dominated by zero values, the OR skip-connection will be performed (Fig.~\ref{mux_0}). Otherwise, the MUX gate will simply keep the output of the second Heaviside (Fig.~\ref{mux_1}). Consequently, this mechanism intrinsically balances zero and one latent values throughout the network, this without any additional regularization.       

\subsection{Fully-quantized LSTM}
LSTM \cite{LSTM} is commonly used because of its capability to capture long-term dependencies within sequences. The basic structure of a cell in LSTM can be described as follows: 

\begin{equation}
i_t = \textrm{sigmoid} \; ( W^i . [x_t, h_{t-1}] + b^i ) 
\label{i_t}
\end{equation}
\begin{equation}
f_t = \textrm{sigmoid} \; ( W^f . [x_t, h_{t-1}] + b^f ) 
\label{f_t}
\end{equation}
\begin{equation}
o_t = \textrm{sigmoid} \; ( W^o . [x_t, h_{t-1}] + b^o ) 
\label{o_t}
\end{equation}
\begin{equation}
\tilde{c}_t =  \textrm{tanh} \; ( W^c . [x_t, h_{t-1}] + b^c ) 
\label{c_tilde}
\end{equation}
\begin{equation}
c_t =  f_t \odot c_{t-1} + i_t \odot \tilde{c}_t
\label{c_t}
\end{equation}
\begin{equation}
h_t =  o_t \odot \textrm{tanh}(c_t)
\label{h_t}
\end{equation}
Eqs.~\ref{i_t}-~\ref{c_tilde} defines the input gate, forget gate, output gate and candidate memory, respectively. Temporal information is transferred along time steps via $c_t$ and $h_t$ (Eqs.~\ref{c_t}-~\ref{h_t}).       

\subsubsection{LSTM weight binarization} \label{qw_lstm}
Let us denote $n_i$, $n_o$ as the dimension of input and output sequences, therefore we have $[x_t, h_{t-1}] \in \mathbb{R}^{n_i + n_o}$. In order to simplify the hardware mapping, all biases are removed. Since the projections will increase the dynamic range of the data, the weight binarization of LSTMs are done with a scaling factor as follows: 
\begin{equation}
\textrm{SSign}(w) =  \frac{3}{\sqrt{n_i + n_o}} \textrm{Sign} (w) 
\label{sbin}
\end{equation}
The scaling factor $\frac{3}{\sqrt{n_i + n_o}}$ is chosen as a compromise between scaling the propagated gradients of the activation functions and matching the bipolar distributions of the later quantized \textbf{sign} and \textbf{heaviside} activations. During backward pass, we still employ the same STE gradient as \cite{BNN}. This scheme is applied to all 4 kernels of the LSTM layers.  

\subsubsection{LSTM activation quantization} \label{LSTM_qact}
\begin{figure}[htbp]
	\centerline{\includegraphics[scale=0.14]{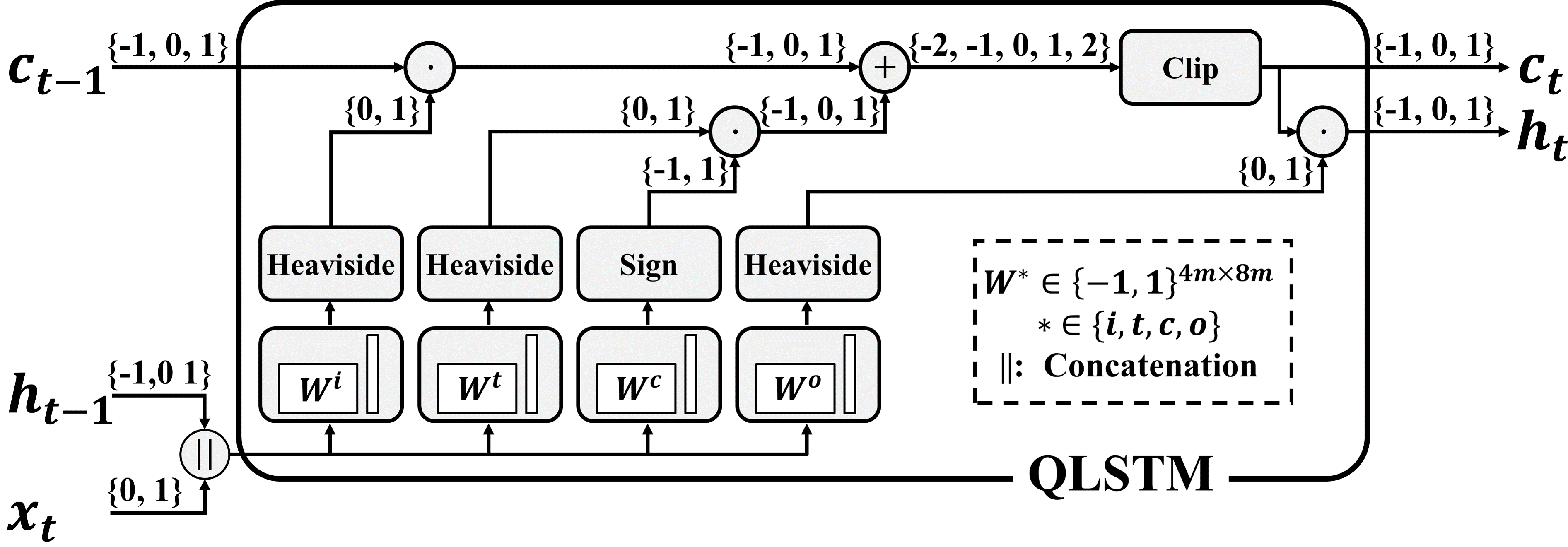}}
	\caption{Computational graph of the proposed Quantized LSTM.}
	\label{QLSTM_graph}
\end{figure}

Whereas quantizing weights is almost straightforward, it becomes much more complex in the case of activations in LSTM due to its internal structure. We replace all \textbf{sigmoid} activations in Eqs.~\ref{i_t}-~\ref{o_t} by the Heaviside function like in~\ref{MOR} and \textbf{tanh} in Eq.~\ref{c_tilde} by a strict Sign. Since the addition in Eq.~\ref{c_t} will increase the dynamic range of data, we will keep $c_t$ values within $\{-1, 0, +1\}$ by using the already introduced Clipped Idendity (Eq.~\ref{clipping}). Consequently, the \textbf{tanh} activation applied to the ternary $c_t$ in Eq.~\ref{h_t} is simply removed, which allows obtaining the output $h_t$ in a ternary representation $\{-1, 0, 1\}$. Figure~\ref{QLSTM_graph} depicts the computational graph of the proposed Quantized LSTM (QLSTM) according to the aforementioned scheme. To better visualize the quantization aspect, we also display the dynamic of internal intermediate values along the connection lines.  

\section{Multi-stage quantization training algorithm} \label{MSQAT}
This work targets a fully-quantized model, including weights, quantizations, hidden states of LSTM and even the Batch Normalizations (BN \cite{BN}). In order to limit the model performance degradation, we apply a multi-stage training procedure in which, we iteratively replace elements of BILLNET by its corresponding quantized version, intrinsically fine-tuning the model to retain the accuracy.

\textbf{$1^{st}$ stage: Training full-precision model.} We firstly train the 32-bit model with ReLU activations in Conv3D part and use it as a proper model initialization.

\textbf{$2^{nd}$ stage: Quantizing all weights.} We keep the full-precision activations and binarize weights using Sign \cite{BNN} for the Conv3D part and SSign for the LSTMs (\cf ~\ref{qw_lstm}). Similarly, for the last Dense layer which reduces the dimension of data from $4m$ to $\#$classes, we also apply a Scaled Ternarization (STern) to its weights: 
\begin{equation}
\textrm{STern}(w) =  \frac{1}{\sqrt{4m}} \textrm{Tern} (w) 
\label{sbin}
\end{equation}
where the Tern function is originally introduced in \cite{TWN}. It is worth mentioning that in hardware implementation, we can simply get rid of these scaling factors, since they do not affect the results of the later Sign/ Heaviside activations and the Argmax operations.

\textbf{$3^{rd}$ stage: Quantizing Conv3D activations.} In this stage, we replace all ReLU by the Heaviside activations while keeping the LSTM activations at full-precision. 

\textbf{$4^{th}$ stage: Removing BNs.} The full-precision affine transform of BN remains an obstacle for model hardware deployment, in particular for 3D CNNs where the data is 4D tensors with an additional temporal dimension. Therefore, we approximate the scaling factors of BN layers in a power-of-2 fashion, which advantageously corresponds to the bitshift operation. Denoting $\mu, \sigma^2$ as the moving mean and the moving variance of BN after the second stage, at inference time, the BN processes the input $x$ to provide the output $y$ as follows:

\begin{equation}
y = \gamma \frac{x - \mu}{\sqrt{\sigma^2 + \epsilon}} + \beta  \equiv \hat{\gamma} x + \hat{\beta}
\label{bn}
\end{equation}
where $\hat{\gamma} = \frac{\gamma}{\sqrt{\sigma^2 + \epsilon}}$ and $\hat{\beta} = \beta -  \frac{\gamma \mu}{\sqrt{\sigma^2 + \epsilon}}$ are equivalent to the scale and the offset vectors. We replace all BN layers by the following offset-free BitShift Normalization (BSN):
\begin{equation}
y = 2^{\lfloor log_2|\hat{\gamma}| \rceil} x 
\end{equation}
Note that in BILLNET, each BN layer is followed by a Heaviside activation. When replacing the BN by BSN, since the equivalent scaling factors are always positive, they will simply keep the sign of data unchanged, therefore, they have no impact to the outcomes of the later Heaviside function. Consequently, all BSNs in BILLNET do not need to be explicitly implemented. 

\textbf{$5^{th}$ stage: Quantizing LSTM activations.} Finally, we replace the \textbf{sigmoid} and \textbf{tanh} used in the LSTM layers (as discussed in subsection ~\ref{LSTM_qact}). The model is now fully quantized with mostly all weights and activations are binarized, except for the ternary output of the QLSTMs and the ternary weights of the last Dense layer.      

\section{Experiments}

\subsection{Settings} 
\textbf{Data pre-processing:} We consider 16-frame sequences with a resolution of $96\times 128$ for training and testing. The Jester Dataset V1 \cite{Jester} is a large-scale hand gesture recognition dataset composed of video clips with a variable number of frames (from 12 to 70). In particular, most of the sequences have between 30-40 frames. Therefore, in order to properly fit the target temporal dimension of 16, we first apply a 2$\times$ temporal down-sampling for sequences of more than 24 frames. This allows us to capture all the hand gestures from end to end. In addition, if the resulting video contains less than 16 frames, we symmetrically repeat the first and the last frames, otherwise, we randomly select the initial time index for the first frame.

\textbf{Training stages:} To train the models, we employ the Adam optimizer \cite{Adam} and the standard Categorical Cross-Entropy (CCE) loss, with a mini-batch size of 40. The configuration of (\textit{initial learning rate}, \textit{number of epochs}) at every stage is respectively (0.0005, 100), (0.0003, 80), (0.0003, 80), (0.0002, 80) and ($10^{-6}$, 80). For each stage, the learning rate firstly remains unchanged, before being exponentially decayed during the last 50 epochs, with a fixed decay rate of $0.85$.   

\textbf{Hardware-related metrics:} We measure the model hardware efficiency in terms of the memory cost using weight-related memory (model size), and the computational complexity using Bit-OPerations (BOPs \cite{BOP}). This allows us to assess the number of parameters and MACs along with the precision of weights and activations. Conventionally, we assume that each full-precision weight and activation requires 32 bits.

\subsection{Results}
\begin{figure}[h]
     \centering
         \begin{subfigure}[b]{0.22\textwidth}
        \includegraphics[width=\textwidth]{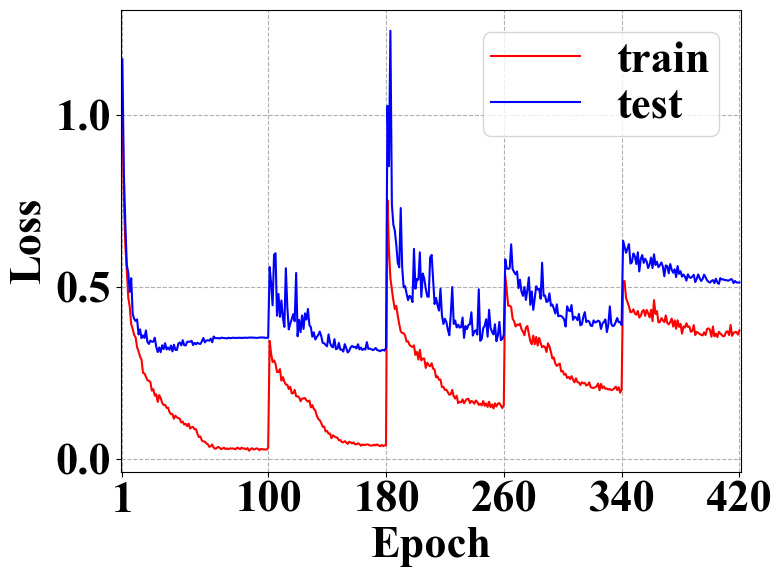}
        \caption{CCE loss.}
        \label{loss}
    \end{subfigure}
    \hspace{-2mm}
    \begin{subfigure}[b]{0.23\textwidth}
        \includegraphics[width=\textwidth]{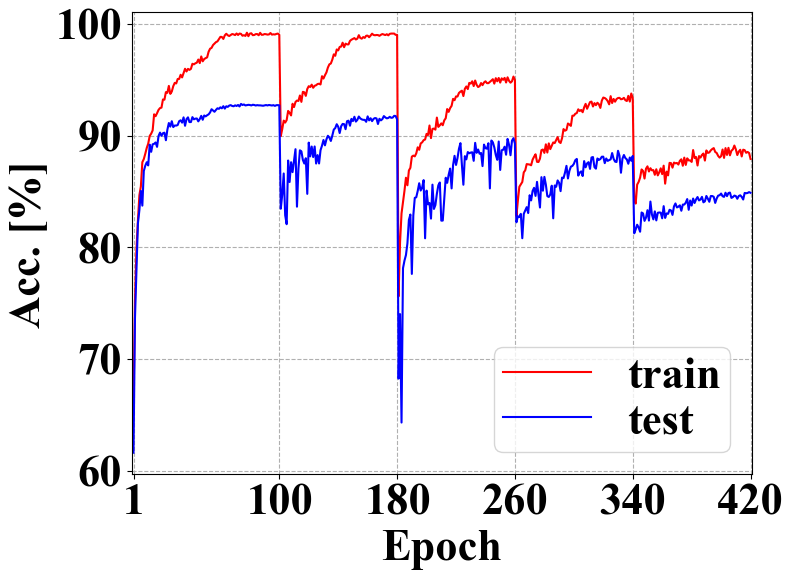}
        \caption{Accuracy metric (Acc.).}
        \label{acc}
    \end{subfigure}
     \caption{Training curves (CCE loss and accuracy) of BILLNET $2\times$ throughout all 5 training stages.}
     \label{training_curves}
\end{figure}

\begin{figure}[h]
     \centering
         \begin{subfigure}[b]{0.175\textwidth}
        \includegraphics[width=\textwidth]{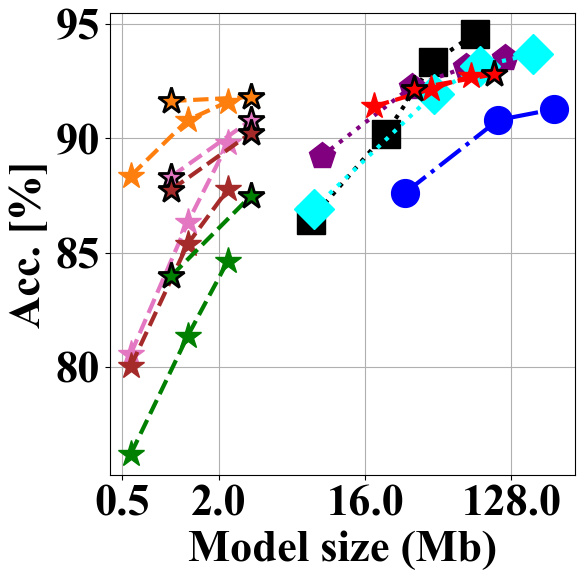}
        \caption{Memory vs. Acc.}
        \label{memory}
    \end{subfigure}
    \hspace{-2mm}
    \begin{subfigure}[b]{0.175\textwidth}
        \includegraphics[width=\textwidth]{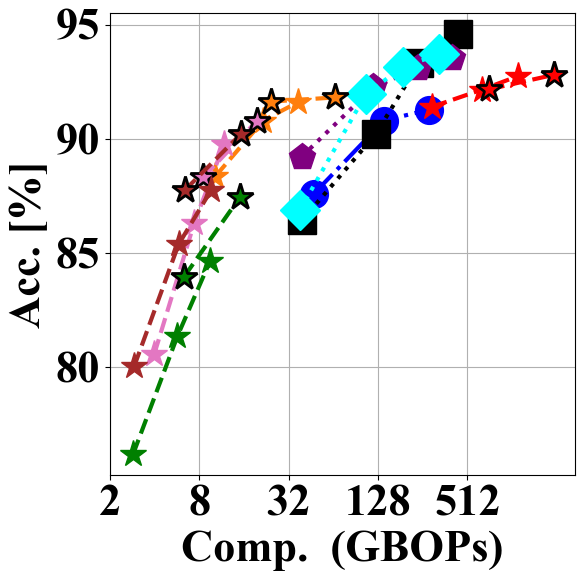}
        \caption{Comp. cost vs. Acc.}
        \label{bop}
    \end{subfigure}
    \hspace{-2mm}
    \begin{subfigure}[b]{0.125\textwidth}
        \includegraphics[width=\textwidth]{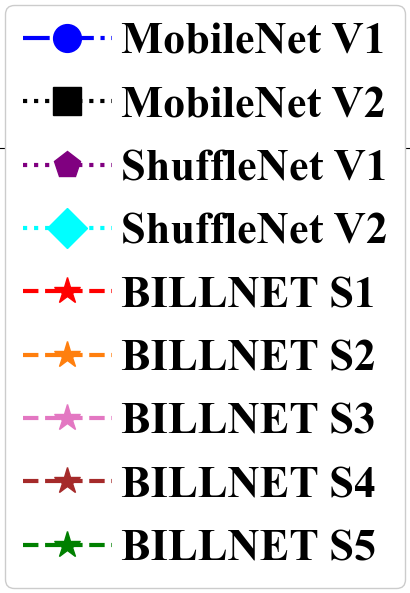}
        \label{legend}
    \end{subfigure}
     \caption{Weight memory (Mb) and computational costs (GBOPs $\sim$ $10^9$ BOPs \cite{BOP}) versus Top-1 Accuracy. Edge-less stars are for BILLNET with $g$=$2$ and $m$=$n$, edged stars are for $m$=$n/2$ and $g=\frac{n}{16}$ (with $n\in \{64, 128\}$).}
     \label{acc_curves}
\end{figure}

\begin{figure}[htbp]
	\centerline{\includegraphics[scale=0.24]{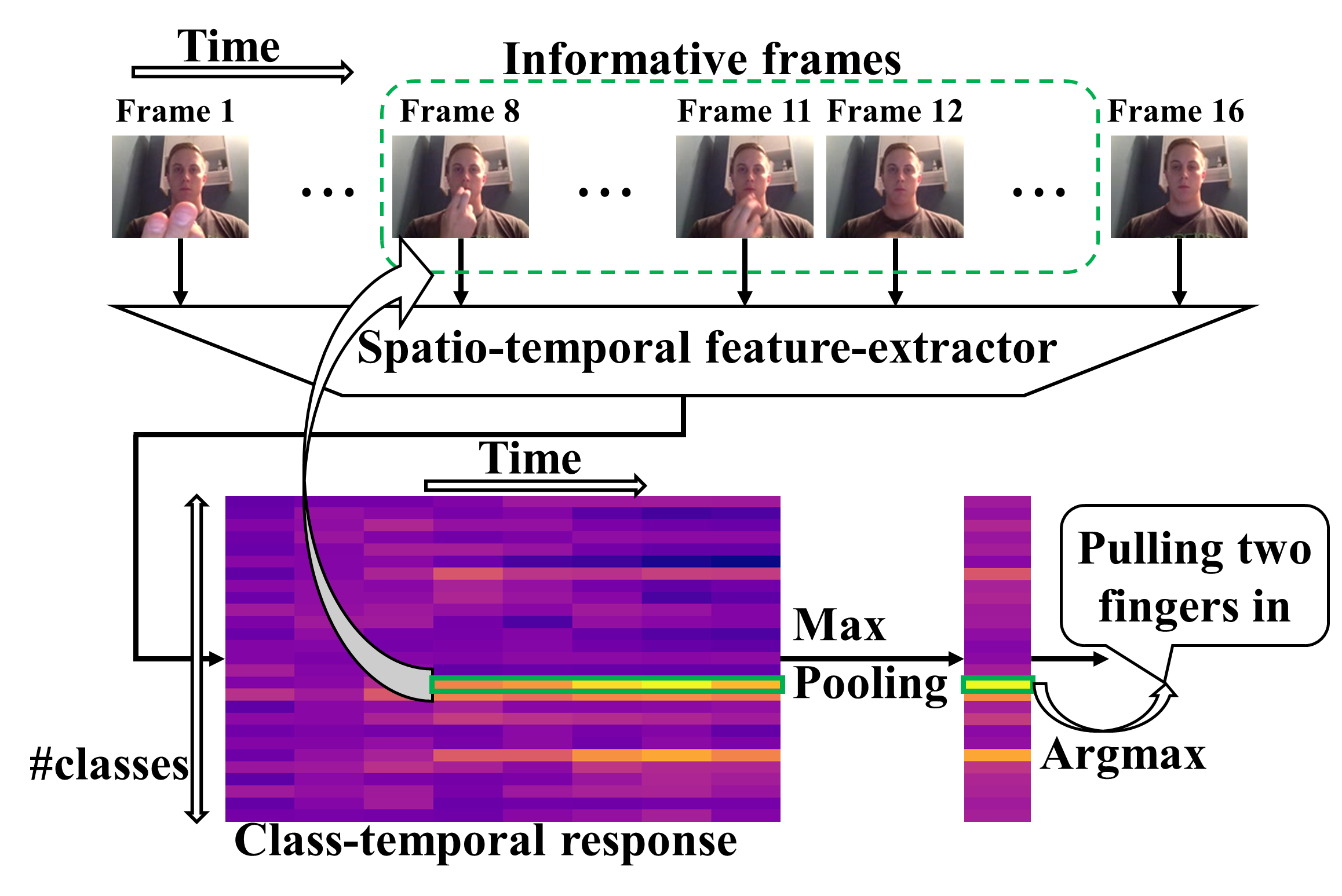}}
	\caption{Class-temporal BILLNET output responses for a Jester test sample labeled as a \textit{Pulling Two Fingers In} gesture. Highlighted time/class positions with maximum values correlate to the most informative frames of the input video. Besides, similar classes (\textit{Pulling Hand In}, \textit{Sliding Two Fingers In}) also exhibit high values at the same columns.}
	\label{heatmap}
\end{figure}

We denote the proposed model with $n=32k$ as BILLNET $k\times$. Figure~\ref{training_curves} reports the evolution of the accuracy and train/test losses. Each training stage (denoted from S1 to S5) allows retaining the accuracy despite introducing quantization effects, even if the remaining gap is significant when quantizing the LSTM activations (S5). Figure~\ref{acc_curves} depicts the accuracy/efficiency compromise of BILLNET and 3D efficient models from \cite{RE3DCNN}. Since \cite{RE3DCNN} does not show the model size and the number of GBOPs for Jester dataset, we compute these values based on their trained models and code (publicly available \footnote{https://github.com/okankop/Efficient-3DCNNs}). It is clear that all quantized versions (S2 to S5) of BILLNET stay on the optimal top-left corner, enabling various types of hardware/accuracy compromises. Table~\ref{results} reports the performance of BILLNET $2\times$ with the specific configuration of $g$=$4$ and $m$=$n/2$ compared to other resource-efficient models. Please note that since BILLNET does not reduce the temporal dimension (except for the first convolution) in order to cap the inference output latency, the full-precision (S1) model involves a higher computational cost than MobileNet and ShuffleNet versions. However, when quantizing the weights and activations, we can advantageously reduce the weight-related memory and computational complexity with at least one order of magnitude. In particular, compared to a 3D-MobileNet V1, the weight-quantized BILLNET $2\times$ (S2) provides a higher accuracy (+$0.8\%$) with a smaller model size ($1\%$) and lower computation needs in terms of \#GBOPs ($17\%$). Besides, GBOP reduction between S3 and S4 ($8.53 \rightarrow 6.39$ GBOPs) shows that it is crucial to replace the BNs by bitshifts, to fully benefit from a hardware simplification in practice. The quantization of LSTM activations has limited impacts on the total number of GBOPs while significantly decreasing the performance from S4 to S5 ($3.78\%$ loss). However, it is still highly relevant considering a dedicated hardware mapping, designed to handle only bit-wise and bit-count operations. Figure~\ref{heatmap} exhibits a class-temporal response of size $8\times27$ for a \textit{Pulling Two Fingers In} gesture example.

\begin{table}[h]
\centering
\caption{Comparison of resource-efficient models on Jester hand gesture dataset, weight-related memory (model size), and bitwidth-aware computational complexity (GBOPs). Results reported here are for BILLNET with $g$=$4$, $n$=$64$ and $m$=$32$.}
\begin{tabular}{| c | c | c | c |}
\hline
\small Model & \small Model size (Mb) & \small Comp. (GBOP)  & \small Acc. ($\%$)  \\
\hline
\small 3D-ShuffleNet V1 \cite{RE3DCNN} & 31.04  & 119.25  & 92.27  \\
\hline
\small 3D-ShuffleNet V2 \cite{RE3DCNN} & 42.56 & 106.09  & 91.96  \\
\hline
\small 3D-MobileNet V1 \cite{RE3DCNN} & 106.56 & 141.02  & 90.81   \\
\hline
\small 3D-MobileNet V2 \cite{RE3DCNN} & 42.24 & 243.18    &  93.34 \\
\hline 
\small BILLNET 2$\times$ -S1 &  32.23   & 718.89    &  92.18    \\
\hline
\small BILLNET 2$\times$ -S2&  1.01   &  24.54   & 91.64    \\
\hline
\small BILLNET 2$\times$ -S3&  1.01   &  8.53   &  88.33   \\
\hline
\small BILLNET 2$\times$ -S4&  1.01   &  6.39  &  87.75  \\
\hline
\small BILLNET 2$\times$ -S5&  1.01   &  6.34  &  83.97  \\
\hline
\end{tabular}
\label{results}
\end{table}

\section{Conclusion}
We introduced a hardware-tiny model for video inference called BILLNET, which involves binary/ternary weights and activations. BILLNET integrates 3D MUX-OR skip connections and a Conv3D factorization to limit the memory and computation needs. Thanks to a 5-stage training procedure, BILLNET offers different hardware-algorithmic trade-offs with significantly reduced model size and \#GBOPs (together with its inherent arithmetic simplifications), while providing an on-par accuracy compared to previously published compact models. More importantly, we aim at designing a hardware-compliant network suitable for later implementation on FPGA or ASIC-based platforms. For this purpose, the fully-quantized BILLNET (S5) can fit a hardware supporting only bit-wise and bit-count operations. Our future works is to revise the last training stage approach (\eg, gradually sharpening the activations during training) to reduce the accuracy loss due to LSTM activations quantization.

\bibliographystyle{IEEEtran}

\begin{thebibliography}{10}
\providecommand{\url}[1]{#1}
\csname url@samestyle\endcsname
\providecommand{\newblock}{\relax}
\providecommand{\bibinfo}[2]{#2}
\providecommand{\BIBentrySTDinterwordspacing}{\spaceskip=0pt\relax}
\providecommand{\BIBentryALTinterwordstretchfactor}{4}
\providecommand{\BIBentryALTinterwordspacing}{\spaceskip=\fontdimen2\font plus
\BIBentryALTinterwordstretchfactor\fontdimen3\font minus
  \fontdimen4\font\relax}
\providecommand{\BIBforeignlanguage}[2]{{%
\expandafter\ifx\csname l@#1\endcsname\relax
\typeout{** WARNING: IEEEtran.bst: No hyphenation pattern has been}%
\typeout{** loaded for the language `#1'. Using the pattern for}%
\typeout{** the default language instead.}%
\else
\language=\csname l@#1\endcsname
\fi
#2}}
\providecommand{\BIBdecl}{\relax}
\BIBdecl

\bibitem{GSN}
S.~Sudhakaran, S.~Escalera, and O.~Lanz, ``Gate-shift networks for video action
  recognition,'' in \emph{2020 IEEE/CVF Conference on Computer Vision and
  Pattern Recognition (CVPR)}, 2020, pp. 1099--1108.

\bibitem{TSN}
L.~Wang, Y.~Xiong, Z.~Wang, Y.~Qiao, D.~Lin, X.~Tang, and L.~Van~Gool,
  ``Temporal segment networks for action recognition in videos,'' \emph{IEEE
  Transactions on Pattern Analysis and Machine Intelligence}, vol.~41, no.~11,
  pp. 2740--2755, 2019.

\bibitem{Gating_revisited}
M.~O. Turkoglu, S.~D'Aronco, J.~Wegner, and K.~Schindler, ``Gating revisited:
  Deep multi-layer {RNNs} that can be trained,'' \emph{IEEE Transactions on
  Pattern Analysis and Machine Intelligence}, pp. 1--1, 2021.

\bibitem{Kinetics}
J.~Carreira and A.~Zisserman, ``Quo vadis, action recognition? a new model and
  the {Kinetics} dataset,'' in \emph{2017 IEEE Conference on Computer Vision
  and Pattern Recognition (CVPR)}, 2017, pp. 4724--4733.

\bibitem{Utube8M}
S.~Abu-El-Haija, N.~Kothari, J.~Lee, A.~Natsev, G.~Toderici, B.~Varadarajan,
  and S.~Vijayanarasimhan, ``{YouTube-8M}: A large-scale video classification
  benchmark,'' \emph{ArXiv}, vol. abs/1609.08675, 2016.

\bibitem{Jester}
J.~Materzynska, G.~Berger, I.~Bax, and R.~Memisevic, ``The {Jester} dataset: A
  large-scale video dataset of human gestures,'' in \emph{2019 IEEE/CVF
  International Conference on Computer Vision Workshop (ICCVW)}, 2019, pp.
  2874--2882.

\bibitem{C3D}
K.~Hara, H.~Kataoka, and Y.~Satoh, ``Can spatiotemporal {3D} cnns retrace the
  history of {2D} cnns and imagenet?'' in \emph{2018 IEEE/CVF Conference on
  Computer Vision and Pattern Recognition}, 2018, pp. 6546--6555.

\bibitem{LSTM}
S.~Hochreiter and J.~Schmidhuber, ``{Long Short-Term Memory},'' \emph{Neural
  Comput.}, vol.~9, no.~8, p. 1735–1780, nov 1997.

\bibitem{RE3DCNN}
O.~Köpüklü, N.~Kose, A.~Gunduz, and G.~Rigoll, ``Resource efficient {3D}
  convolutional neural networks,'' in \emph{2019 IEEE/CVF International
  Conference on Computer Vision Workshop (ICCVW)}, 2019, pp. 1910--1919.

\bibitem{TinyVN}
A.~J. Piergiovanni, A.~Angelova, and M.~S. Ryoo, ``Tiny video networks,''
  \emph{ArXiv}, vol. abs/1910.06961, 2019.

\bibitem{3D-FPGA-Pruning}
M.~Sun, P.~Zhao, M.~Gungor, M.~Pedram, M.~Leeser, and X.~Lin, ``{3D} {CNN}
  acceleration on {FPGA} using hardware-aware pruning,'' in \emph{2020 57th
  ACM/IEEE Design Automation Conference (DAC)}, 2020, pp. 1--6.

\bibitem{QNN}
I.~Hubara, M.~Courbariaux, D.~Soudry, R.~El-Yaniv, and Y.~Bengio, ``Quantized
  neural networks: Training neural networks with low precision weights and
  activations,'' \emph{J. Mach. Learn. Res.}, vol.~18, no.~1, p. 6869–6898,
  jan 2017.

\bibitem{BNN}
I.~Hubara, M.~Courbariaux, D.~Soudry, R.~El{-}Yaniv, and Y.~Bengio, ``Binarized
  neural networks,'' in \emph{Advances in Neural Information Processing Systems
  (NeurIPS)}, 2016, pp. 4107--4115.

\bibitem{LSQ}
S.~K. Esser, J.~L. McKinstry, D.~Bablani, R.~Appuswamy, and D.~S. Modha,
  ``Learned step size quantization,'' in \emph{8th International Conference on
  Learning Representations, {ICLR} 2020, Addis Ababa, Ethiopia, April 26-30,
  2020}, 2020.

\bibitem{RNN_low_prec}
J.~Ott, Z.~Lin, Y.~Zhang, S.-C. Liu, and Y.~Bengio, ``Recurrent neural networks
  with limited numerical precision,'' \emph{ArXiv}, vol. abs/1611.07065, 2016.

\bibitem{EQRNN}
M.~Z. Alom, A.~T. Moody, N.~Maruyama, B.~C. Van~Essen, and T.~M. Taha,
  ``Effective quantization approaches for recurrent neural networks,'' in
  \emph{2018 International Joint Conference on Neural Networks (IJCNN)}, 2018,
  pp. 1--8.

\bibitem{ResNet}
K.~He, X.~Zhang, S.~Ren, and J.~Sun, ``Deep residual learning for image
  recognition,'' in \emph{2016 IEEE Conference on Computer Vision and Pattern
  Recognition (CVPR)}, 2016, pp. 770--778.

\bibitem{ResAttention}
F.~Wang, M.~Jiang, C.~Qian, S.~Yang, C.~Li, H.~Zhang, X.~Wang, and X.~Tang,
  ``Residual attention network for image classification,'' in \emph{2017 IEEE
  Conference on Computer Vision and Pattern Recognition (CVPR)}, 2017, pp.
  6450--6458.

\bibitem{SeNet}
J.~Hu, L.~Shen, S.~Albanie, G.~Sun, and E.~Wu, ``Squeeze-and-excitation
  networks,'' \emph{IEEE Transactions on Pattern Analysis and Machine
  Intelligence}, vol.~42, no.~8, pp. 2011--2023, 2020.

\bibitem{3DResNet}
K.~Hara, H.~Kataoka, and Y.~Satoh, ``Learning spatio-temporal features with
  {3D} residual networks for action recognition,'' in \emph{2017 IEEE
  International Conference on Computer Vision Workshops (ICCVW)}, 2017, pp.
  3154--3160.

\bibitem{Revisiting3RN}
X.~Du, Y.~Li, Y.~Cui, R.~Qian, J.~Li, and I.~Bello, ``Revisiting {3D} resnets
  for video recognition,'' \emph{ArXiv}, vol. abs/2109.01696, 2021.

\bibitem{ST_Attention}
J.~Li, X.~Liu, W.~Zhang, M.~Zhang, J.~Song, and N.~Sebe, ``Spatio-temporal
  attention networks for action recognition and detection,'' \emph{IEEE
  Transactions on Multimedia}, vol.~22, no.~11, pp. 2990--3001, 2020.

\bibitem{MixedConv}
D.~Tran, H.~Wang, L.~Torresani, J.~Ray, Y.~LeCun, and M.~Paluri, ``A closer
  look at spatiotemporal convolutions for action recognition,'' in \emph{2018
  IEEE/CVF Conference on Computer Vision and Pattern Recognition}, 2018, pp.
  6450--6459.

\bibitem{RethinkingConv3D_byMixedConv}
S.~Xie, C.~Sun, J.~Huang, Z.~Tu, and K.~Murphy, ``Rethinking spatio-temporal
  feature learning: Speed-accuracy trade-offs in video classification,'' in
  \emph{Computer Vision - {ECCV} 2018 - 15th European Conference, Munich,
  Germany, September 8-14, 2018, Proceedings, Part {XV}}, V.~Ferrari,
  M.~Hebert, C.~Sminchisescu, and Y.~Weiss, Eds., pp. 318--335.

\bibitem{P3D}
Z.~Qiu, T.~Yao, and T.~Mei, ``Learning spatio-temporal representation with
  pseudo-{3D} residual networks,'' in \emph{2017 IEEE International Conference
  on Computer Vision (ICCV)}, 2017, pp. 5534--5542.

\bibitem{TSM}
J.~Lin, C.~Gan, and S.~Han, ``{TSM}: Temporal shift module for efficient video
  understanding,'' in \emph{2019 IEEE/CVF International Conference on Computer
  Vision (ICCV)}, 2019, pp. 7082--7092.

\bibitem{MobileNet}
\BIBentryALTinterwordspacing
A.~G. Howard, M.~Zhu, B.~Chen, D.~Kalenichenko, W.~Wang, T.~Weyand,
  M.~Andreetto, and H.~Adam, ``Mobilenets: Efficient convolutional neural
  networks for mobile vision applications,'' \emph{CoRR}, vol. abs/1704.04861,
  2017. [Online]. Available: \url{http://arxiv.org/abs/1704.04861}
\BIBentrySTDinterwordspacing

\bibitem{ShuffleNet}
X.~Zhang, X.~Zhou, M.~Lin, and J.~Sun, ``Shufflenet: An extremely efficient
  convolutional neural network for mobile devices,'' in \emph{2018 IEEE/CVF
  Conference on Computer Vision and Pattern Recognition}, 2018, pp. 6848--6856.

\bibitem{DynamicQ}
X.~Sun, R.~Panda, C.-F.~R. Chen, A.~Oliva, R.~Feris, and K.~Saenko, ``Dynamic
  network quantization for efficient video inference,'' in \emph{Proceedings of
  the IEEE/CVF International Conference on Computer Vision (ICCV)}, October
  2021, pp. 7375--7385.

\bibitem{Bin3DCNN}
G.~Li, M.~Zhang, Q.~Zhang, and Z.~Lin, ``Efficient binary {3D} convolutional
  neural network and hardware accelerator,'' \emph{J. Real Time Image
  Process.}, vol.~19, no.~1, pp. 61--71, 2022.

\bibitem{4bQLSTM}
A.~Fasoli, C.~Chen, M.~J. Serrano, X.~Sun, N.~Wang, S.~Venkataramani, G.~Saon,
  X.~Cui, B.~Kingsbury, W.~Zhang, Z.~T{\"{u}}ske, and K.~Gopalakrishnan,
  ``4-bit quantization of {LSTM}-based speech recognition models,''
  \emph{CoRR}, vol. abs/2108.12074, 2021.

\bibitem{HitNet}
P.~Wang, X.~Xie, L.~Deng, G.~Li, D.~Wang, and Y.~Xie, ``{HitNet}: Hybrid
  ternary recurrent neural network,'' in \emph{Advances in Neural Information
  Processing Systems (NeurIPS)}, vol.~31, 2018.

\bibitem{DCLSTM}
G.~Nan, C.~Wang, W.~Liu, and F.~Lombardi, ``{DC-LSTM}: Deep compressed {LSTM}
  with low bit-width and structured matrices,'' in \emph{2020 IEEE
  International Symposium on Circuits and Systems (ISCAS)}, 2020, pp. 1--5.

\bibitem{STE}
Y.~Bengio, N.~Léonard, and A.~Courville, ``Estimating or propagating gradients
  through stochastic neurons for conditional computation,''
  \emph{arXiv:1308.3432 [cs]}, Aug. 2013.

\bibitem{BN}
S.~Ioffe and C.~Szegedy, ``Batch normalization: Accelerating deep network
  training by reducing internal covariate shift,'' \emph{ArXiv}, vol.
  abs/1502.03167, 2015.

\bibitem{TWN}
\BIBentryALTinterwordspacing
F.~Li and B.~Liu, ``Ternary weight networks,'' \emph{CoRR}, vol.
  abs/1605.04711, 2016. [Online]. Available:
  \url{http://arxiv.org/abs/1605.04711}
\BIBentrySTDinterwordspacing

\bibitem{Adam}
D.~P. Kingma and J.~Ba, ``Adam: {A} method for stochastic optimization,'' in
  \emph{3rd International Conference on Learning Representations, {ICLR} 2015,
  San Diego, CA, USA, May 7-9, 2015, Conference Track Proceedings}, Y.~Bengio
  and Y.~LeCun, Eds., 2015.

\bibitem{BOP}
Y.~Wang, Y.~Lu, and T.~Blankevoort, ``Differentiable joint pruning and
  quantization for hardware efficiency,'' in \emph{Computer Vision - {ECCV}
  2020 - 16th European Conference, Glasgow, UK, August 23-28, 2020,
  Proceedings, Part {XXIX}}, ser. Lecture Notes in Computer Science,
  A.~Vedaldi, H.~Bischof, T.~Brox, and J.~Frahm, Eds., vol. 12374.\hskip 1em
  plus 0.5em minus 0.4em\relax Springer, pp. 259--277.

\end{thebibliography}

\end{document}